\documentclass[10pt,twocolumn,letterpaper]{article}

\usepackage[pagenumbers]{cvpr}
\usepackage{cvpr}
\usepackage{times}
\usepackage{epsfig}
\usepackage{graphicx}
\usepackage{amsmath}
\usepackage{amssymb}
\usepackage{multirow}

\usepackage[accsupp]{axessibility}

\newcommand{\tabincell}[2]{\begin{tabular}{@{}#1@{}}#2\end{tabular}}

\begin{document}

\title{SynFacePAD 2023: Competition on Face Presentation Attack Detection Based on Privacy-aware Synthetic Training Data
\vspace{-3mm}}

\author{Meiling Fang$^{1,2,*}$, Marco Huber$^{1,2,*}$, Julian Fierrez$^{3,*}$, Raghavendra Ramachandra$^{4,*}$, Naser Damer$^{1,2,*}$, \\ 
Alhasan Alkhaddour$^{5,+}$, Maksim Kasantcev$^{5,+}$, Vasiliy Pryadchenko$^{5,+}$,  
Ziyuan Yang$^{6,+}$, \\ 
Huijie Huangfu$^{6,+}$, Yingyu Chen$^{6,+}$, Yi Zhang$^{6,+}$,  Yuchen Pan$^{7,+}$, Junjun Jiang$^{7,+}$,  \\ 
Xianming Liu$^{7,+}$, 
Xianyun Sun$^{8,+}$, Caiyong Wang$^{8,+}$, Xingyu Liu$^{8,+}$, Zhaohua Chang$^{8,+}$, \\
Guangzhe Zhao$^{8,+}$, 
Juan Tapia$^{9,10,+}$, Lazaro Gonzalez-Soler$^{9,+}$, Carlos Aravena$^{10,+}$, Daniel Schulz$^{10,+}$ \\
$^{1}$Fraunhofer Institute for Computer Graphics Research IGD,
Darmstadt, Germany\\
$^{2}$Department of Computer Science, TU Darmstadt,
Darmstadt, Germany\\
$^{3}$Biometrics and Data Pattern Analytics Lab, Universidad Autonoma de Madrid, Spain\\
$^{4}$Norwegian University of Science and Technology (NTNU), Norway\\
$^{5}$ID R\&D, Inc., New York, US\\
$^{6}$School of Cyber Science and Engineering, Sichuan University, Chengdu, China \\
$^{7}$School of Computer Science and Technology, Harbin Institute of Technology, Harbin, China\\
$^{8}$School of Electrical and Information Engineering, \\ Beijing University of Civil Engineering and Architecture, China\\
$^{9}$Biometrics and Security Research Group, Hochschule Darmstadt, Darmstadt, Germany \\
$^{10}$I+D Vision Center, Santiago, Chile \\
$^{*}$Competition Organizers. 
$^{+}$Competition participant. \\
Email: {meiling.fang@igd.fraunhofer.de}
\vspace{-7mm}
}

\maketitle
\thispagestyle{empty}

\begin{abstract}
\vspace{-3mm}
This paper presents a summary of the Competition on Face Presentation Attack Detection Based on Privacy-aware Synthetic Training Data (SynFacePAD 2023) held at the 2023 International Joint Conference on Biometrics (IJCB 2023). The competition attracted a total of 8 participating teams with valid submissions from academia and industry. The competition aimed to motivate and attract solutions that target detecting face presentation attacks while considering synthetic-based training data motivated by privacy, legal and ethical concerns associated with personal data. To achieve that, the training data used by the participants was limited to synthetic data provided by the organizers. The submitted solutions presented innovations and novel approaches that led to outperforming the considered baseline in the investigated benchmarks.
\vspace{-7mm}
\end{abstract}

\section{Introduction} 
Face recognition has been widely deployed in various application scenarios, such as access control, phone unlocking, and mobile payment. Reasons for this include its convenience and outstanding performance \cite{DBLP:conf/cvpr/DengGXZ19,DBLP:conf/icb/BoutrosDFKK21,elasticface}.
However, face recognition is susceptible to Presentation Attacks (PAs), such as high-resolution photos and videos of an authorized user \cite{casia_fas,replay_attack,oulu_npu,DBLP:journals/pr/FangDKK22,2023_COMPSAC_PAD-Phys}. 
Therefore, face Presentation Attack Detection (PAD) technology \cite{2023_Book-PAD_Face}, which describes the process of identifying whether a face presented to the system is a bona fide (live) or PA,  plays an important role to secure recognition from PAs \cite{hadid15SPMspoofing}. These PA detectors are often built using authentic biometric data \cite{galbally14TIP}, raising ethical and legal challenges. Such challenges have recently been discussed in face recognition \cite{BOUTROS2023104688,DBLP:conf/iccv/QiuYG00T21}, face morphing attack detection \cite{DBLP:conf/cvpr/DamerLFSPB22,2022_Handbook_Trends_RT}, and face PAD \cite{synthaspoof}. There was previously a series of competitions on face PAD based on authentic data \cite{DBLP:conf/icb/PurnapatraSBDYM21} and a competition targeting face morphing attack detection based on privacy synthetic training data \cite{DBLP:conf/icb/HuberBLRRDNGSCT22}. However, this is the first competition targeting PAs on face recognition while limiting its development data to synthetic data.
Given the legal privacy regulations, the collection, use, share, and maintenance of face data for biometric processing is extremely challenging \cite{onsyntheticdata_20220922}. For example, several large-scale face recognition datasets \cite{DBLP:conf/fgr/CaoSXPZ18,DBLP:conf/eccv/GuoZHHG16,DBLP:conf/cvpr/Kemelmacher-Shlizerman16} were withdrawn by their creators with privacy and proper subjects consent issues being the main reason. 
One of the main solutions for this issue is the use of synthetic data \cite{onsyntheticdata_20220922}. This has been very recently and successfully proposed for the training of face recognition \cite{DBLP:conf/iccv/QiuYG00T21,DBLP:conf/icb/BoutrosHSRD22,DBLP:conf/fgr/BoutrosKFKD23} and morphing attack detection \cite{DBLP:conf/cvpr/DamerLFSPB22,DBLP:conf/icb/HuberBLRRDNGSCT22,DBLP:conf/icb/FangBD22,DBLP:conf/iwbf/DamerFSKHB23}, among other processes such as model quantization \cite{DBLP:conf/icpr/BoutrosDK22,DBLP:journals/ivc/KolfEB0D23}. Furthermore, a recent work followed this motivation to take advantage of synthetic data to develop PADs in a privacy-friendly manner \cite{synthaspoof} and proved the usability of synthetic data for the development of face PADs. The utilized assumption is that learning to detect the differences between bona fide and attack samples of a synthetic origin can be used to detect these differences between authentic bona fide and attacks and thus train PAD without authentic private data.

Driven by the need for the development of face PAD datasets that prioritize the privacy of individuals, promote data sharing within the research community, and ensure the reproducibility and continuity of face PAD research, we conduct the SynFacePAD 2023: Competition on Face Presentation Attack Detection Based on Privacy-aware Synthetic Training Data at the International Joint Conference on Biometrics 2023. The results and observations are summarized in this paper.

\section{Dataset, Evaluation Criteria, and Participants} 
\label{sec:comp_info}
\vspace{-5mm}
\begin{figure*}[htbp]
\centering
\includegraphics[width=\linewidth]{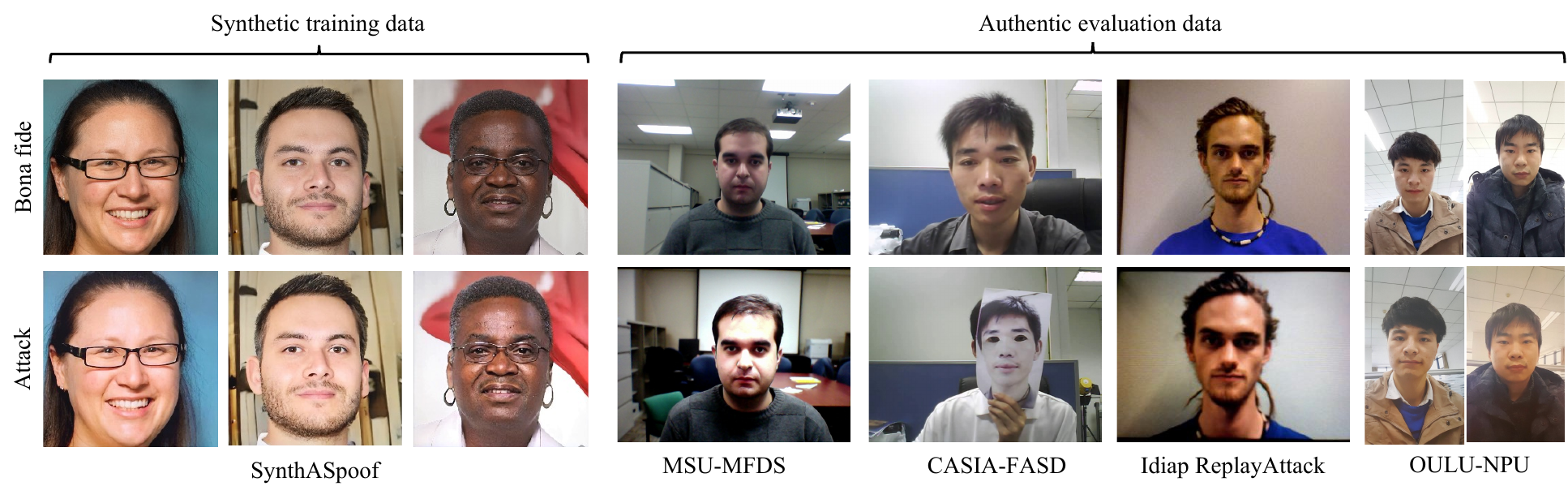}
\caption{Samples of the synthetic training data from SynthASpoof \cite{synthaspoof}, as well as four authentic evaluation face PAD benchmarks.}
\label{fig:image_samples}
\vspace{-2mm}
\end{figure*}

\begin{table}[htb]
\centering
\resizebox{0.47\textwidth}{!}{
\begin{tabular}{l|c|c|c|c}
\hline 
Dataset & Year & \# Bona fide/attack  & \# Sub & Attack types  \\ \hline \hline
SynthASpoof \cite{synthaspoof} & 2023 & 25,000 / 78,800 (I\&V)  & 25,000 & 1 Print, 3 Replay \\ \hline 
CASIA-FASD \cite{casia_fas} & 2012 & 150 / 450 (V) & 50 & 1 Print, 1 Replay \\ 
Replay-Attack \cite{replay_attack} & 2012 & 200 / 1,000 (V) & 50  & 1 Print, 2 Replay   \\ 
MSU-MFSD \cite{msu_mfs} & 2015 & 70 / 210 (V) & 35 & 1 Print, 2 Replay \\
OULU-NPU \cite{oulu_npu}  & 2017 & 1,980 / 3,960 (V) & 55 & 2 Print,2 Replay \\ \hline
\end{tabular}}
\caption{Summary of the used face PAD datasets. V and I are shorthand for video and image, respectively. SynthASpoof is a synthetic face PAD, serving as training dataset, and the other four are public available authentic face PAD evaluation benchmarks.}
\label{tab:datasets-summariz}
\vspace{-5mm}
\end{table}

\subsection{Dataset}
\label{ssec:dataset}
\paragraph{Training Dataset:} To promote the development of face PAD on synthetic data, this competition restricts the training data to the provided privacy-friendly synthetic dataset, \textbf{SynthASpoof} \cite{synthaspoof}. The SynthASpoof consists of 25,000 bona fide and 78,800 attack samples and is publicly available. The bona fide samples were generated using StyleGAN2-ADA \cite{DBLP:conf/nips/KarrasAHLLA20} and the attack samples were collected by presenting these synthetic samples as printed or replayed attacks to three different capture sensors. To ensure that the participants trained their solutions solely by using the training data set provided, the solutions were trained again by the organizers.
\vspace{-4mm}
\paragraph{Evaluation Benchmarks:} For the evaluation, we use four authentic face PAD benchmarks: MSU-MFSD \cite{msu_mfs} (denoted as M), CASIA-MFSD \cite{casia_fas} (denoted as C), Idiap Replay-Attack \cite{replay_attack} (denoted as I), and OULU-NPU \cite{oulu_npu} (denoted as O). We select these four datasets by considering their widely used in generalized face PAD studies \cite{DBLP:conf/icb/FangAKD22,Fang_2022_WACV,DBLP:conf/cvpr/LiPWK18,DBLP:conf/cvpr/ShaoLLY19,DBLP:conf/aaai/ShaoLY20}.
The \textbf{MSU-MFSD} (M) \cite{msu_mfs} dataset comprises 440 videos captured from 35 subjects, utilizing two different resolutions of cameras. The dataset includes two types of attacks: printed photo attacks and replay attacks.
The \textbf{CASIA-MFSD} (C) \cite{casia_fas} dataset consists of 600 videos from 50 subjects and includes three types of attacks: warped photo attack, cut photo attack, and video replay attack.
The \textbf{Idiap Replay-Attack} (I) \cite{replay_attack} dataset contains 300 videos from 50 subjects captured under different sensors and illumination conditions. The dataset includes two types of attacks: print attacks and replay attacks.
The \textbf{Oulu-NPU} (O) \cite{oulu_npu} is a mobile face PAD dataset collected in a realistic mobile scenario. It consists of 5940 video clips from 55 subjects using six different mobile phones. 

Samples from the provided training dataset SynthASpoof, as well as four evaluation benchmarks, are shown in Figure \ref{fig:image_samples}, and the corresponding information is summarized in Table \ref{tab:datasets-summariz}.
In addition, we provide participants with a pre-processing implementation that includes face detection and cropping\footnote{\url{https://github.com/meilfang/SynthASpoof/tree/main/data_preprocess}}. For the evaluation benchmark, the faces were detected and cropped using the MTCNN method. Notably, there are no restrictions on the pre-processing of the training data.

\vspace{-2mm}
\subsection{Baseline Methods}
The baseline performance is based on two face PA detectors reported in \cite{synthaspoof}, ResNet and PixBis. ResNet is is one of the most popular backbone architectures used in face PAD algorithm design \cite{DBLP:journals/tbbis/YuLSXZ21,Fang_2022_WACV,DBLP:conf/eccv/ZhangYLYYSL20,DBLP:conf/icb/FangAKD22}. PixBis \cite{DBLP:conf/icb/GeorgeM19} employs a binary supervisory strategy at pixel-level to simplify the problem and obviate the need for a computationally intensive synthesis of depth maps.

\begin{table*}[ht]
\centering
\resizebox{0.9\textwidth}{!}{
\begin{tabular}{llll|l}
\hline
Team & Team members & Affiliations & Type & Solution \\ \hline \hline
ID R\&D Inc & \tabincell{l}{Alhasan Alkhaddour, Maksim Kasantcev, \\ Vasiliy Pryadchenko} & ID R\&D, Inc., New York, US & Industry & ViT-SIDE B \\ \hline
\multirow{2}{*}{SCU-DIG} & \multirow{2}{*}{\tabincell{l}{Ziyuan Yang, Huijie Huangfu, Yingyu Chen, \\ Yi Zhang}} & \multirow{2}{*}{\tabincell{l}{School of Cyber Science and Engineering,\\Sichuan University, Chengdu, China}} & \multirow{2}{*}{Academic} & SynFace Co-Former B \\ 
 &  &  &  & SynFace Co-Former A \\ \hline
\multirow{2}{*}{HIT} & \multirow{2}{*}{Yuchen Pan, Junjun Jiang, Xianming Liu} & \multirow{2}{*}{\tabincell{l}{School of Computer Science and Technology, \\ Harbin Institute of Technology, Harbin, China}} & \multirow{2}{*}{Academic} & CoDe-Lc \\ 
 &  &  &  & CoDe-Lh \\ \hline
BUCEA & \tabincell{l}{Xianyun Sun, Caiyong Wang, Xingyu Liu, \\ Zhaohua Chang, Guangzhe Zhao} & \tabincell{l}{School of Electrical and Information Engineering, \\ Beijing University of Civil Engineering and Architecture, China} & Academic & OrthPADNet \\ \hline
hda & Juan Tapia, Lazaro J. Gonzalez-Soler & \tabincell{l}{Biometrics and Security Research Group, \\ Hochschule Darmstadt, Darmstadt, Germany} & Academic & hdaFVPAD \\ \hline 
idvc & Juan Tapia, Carlos Aravena, Daniel Schulz & I+D Vision Center, Santiago, Chile & Industry & idvcVT \\ \hline
\multirow{2}{*}{Anonymous-1} & - & - & \multirow{2}{*}{Industry} & Saliency-ResNet-CAS  \\ 
 & - & - &  & Saliency-ResNet-ES  \\ \hline
Anonymous-2 &-  & - & Acadmic & - \\ \hline
\end{tabular}}
\caption{A summary of the valid participating teams, team members, affiliations, type of institution, and solutions. More details on the submitted algorithms is provided in Section \ref{sec:submitted_solutions}.}
\label{tab:partipants_info}
\vspace{-5mm}
\end{table*}

\vspace{-2mm}
\subsection{Evaluation Criteria}
The SynFacePAD competition uses two PAD metrics to evaluate the submitted solutions following the ISO/IEC 30107-3 \cite{ISO301073} standard: Bona fide Presentation Classification Error Rate (BPCER) and Attack Presentation Classification Error Rate (APCER). BPCER refers to the proportion of bona fide presentations classified as attack samples, while APCER is the proportion of attack presentations incorrectly classified as bona fide presentation. The submitted solutions are evaluated at two different fixed APCER (and BPCER) values, 10\% and 20\%, and the corresponding BPCER (and APCER) is reported. To cover diverse operational points and enable a detailed results discussion, we provide a visual evaluation by plotting Receiver Operating Characteristic (ROC) curves, where the x-axis of the ROC is APCER and the y-axis is 1-BPCER. 
Furthermore, following existing cross-domain face PAD methods \cite{synthaspoof,DBLP:conf/cvpr/LiPWK18,DBLP:conf/cvpr/ShaoLLY19,DBLP:conf/aaai/ShaoLY20}, the Half Total Error Rate (HTER), which is the mean of BPCER and APCER \cite{ISO301073} and Area under the ROC Curve (AUC) value is also reported. The HTER threshold is computed based on the Equal Error Rate (EER) threshold from the targeted evaluating benchmark directly.

The ranking of the submitted solutions on each benchmark is determined by the APCER at a fixed BPCER of 20\%, allowing us to analyze the detectability of different solutions on the attack samples. Once the solutions are ranked, the final ranking of the team is based on their best-performing solution if a team submits two solutions.

\vspace{-3mm}
\subsection{Competition Participants}

The goal of the SynFacePAD competition is to attract participants from both academia and industry, with a wide geographic and activity variation. The call for participation was shared on the website of the International Joint Conference on Biometrics (IJCB) 2023, the competition’s own website, and various social media platforms. As a result, 14 registered teams, both from academia and industry registered for the competition. Among them, eight teams submitted a total of 11 valid solutions, with each team allowed to submit up to two solutions. These eight teams have affiliations in five different countries, consisting of five teams with academic affiliations and three teams with industry affiliations. Two teams opted to be anonymous. Table \ref{tab:partipants_info} provides a summary of the participating teams.

\subsection{Submission and Evaluation Process}
Each team participating in the SynFacePAD competition registered with a team name and a list of team members with their affiliations for the competition and was then provided access to the synthetic data. The training data was restricted to the use of the synthetic data provided by the organizers which consisted of the SynthASpoof dataset \cite{synthaspoof}. Only this dataset was allowed to be used during the training of the PADs.
Teams were allowed to use pre-trained weights on non-face data. The organizers provided pre-processing code for the training data, but it was not mandatory for teams to employ it. Each team was then requested to submit either a Win32 or Linux executable or, if wanted, their Python script. To ensure the integrity of the competition, the top three ranked solutions were examined by the organizers, i.e. these models were re-trained to validate that only the provided synthetic data was used for training and no pre-trained weights for faces were used. All solutions were evaluated on a restricted system without an internet connection to prevent any potential data leaks.

\begin{table*}[]
\resizebox{1.0\textwidth}{!}{
\begin{tabular}{lllllclcc}
\hline
Solution & Base architecture & Augmentation & Init. Weight & Loss function & Hardware & Selec. of model & FLOPs (G) & Param. (M) \\ \hline \hline
ViT-SIDE B & ViT-Tiny & \tabincell{l}{CJ, JPEG compression, \\ rotation, blur, random crop} & ImageNet & CE & NVIDIA T4, 16GB & \tabincell{l}{Minmun loss \\ with a maximum \\ epoch of 100 and \\  a patience of 20} & 1.08 & 5.524 \\ \hline
SynFace Co-Former B & \multirow{2}{*}{Swin-transformer} & \tabincell{l}{CJ, HF, SR, GA, \\ RGB-S} & \multirow{2}{*}{scratch} & \multirow{2}{*}{CE} & \multirow{2}{*}{RTX 3090, 24GB} & \multirow{2}{*}{At 6 epochs} & \multirow{2}{*}{28.09} & \multirow{2}{*}{168.74} \\ \cline{1-1} \cline{3-3}
SynFace Co-Former A &  & \tabincell{l}{Proposed reflection \\ simulation augmentation} &  &  &  &  &  &  \\ \hline
CoDe-Lc & \multirow{2}{*}{AlexNet} & CJ, HF, SR, GA, RGB-S & \multirow{2}{*}{scratch} & \tabincell{l}{BCE, MSE, \\ Cosine similarity loss} & \multirow{2}{*}{RTX 3080 Ti, 12GB} & \multirow{2}{*}{At 200 epochs} & 1.42 & 115.06 \\ \cline{1-1} \cline{3-3} \cline{5-5} \cline{8-9} 
CoDe-Lh &  & + Gaussian blur &  & \tabincell{l}{BCE, MSE, \\ hypersphere loss} &  &  & 1.42 & 114.53 \\ \hline
OrthPADNet & ResNet-18 & \tabincell{l}{CJ, HF, SR, GA, \\ RGB-S, JPEG compression} & scratch & \tabincell{l}{CE with \\ orthogonal projection loss, \\ BCE, \\ Cosine similarity loss} & RTX 3090, 24GB & At 30 epochs & 43.81 & 55.4 \\ \hline
hdaFVPAD & Fisher Vector + SVM & - & - & - & - & - & - & 0.0015 \\ \hline
idvcVT & Swin-Transformer & - & ImageNet & BCE & RTX 2080 Ti, 11GB & Grid Search & 4.36 & 28.29 \\ \hline
Saliency-ResNet-CAS & ResNet-50 & \multirow{2}{*}{\tabincell{l}{CJ, random erasing, \\ channel normalization}} & - & \multirow{2}{*}{FocalLoss} & \multirow{2}{*}{RTX 3060, 6GB} & \multirow{2}{*}{Best AUC at val set} & \multirow{2}{*}{-} & \multirow{2}{*}{36.92} \\
Saliency-ResNet-ES & ResNet-50 &  &  &  &  &  &  & \\ \hline
\end{tabular}}
\caption{Basic details of the submitted algorithms. SynFace Co-Former B and A refer to Baseline and the proposed reflection simulation Augmentation technique. Saliency-ResNet-CAS indicates use Cosine Annealing Scheduler (CAS) and Saliency-ResNet-ES refers to use Exponential Scheduler (ES). CJ, HF, SR, GA, RGB-S in Augmentation column refer to Color Jittering, Horizontal Flipping, Scale and Rotation, Gamma Adjustment, and RGB-Shift, respectively. }
\label{tab:solutions_info}
\vspace{-5mm}
\end{table*}

\vspace{-3mm}
\section{Submitted Solutions} 
\label{sec:submitted_solutions}

In total, 14 teams have registered for the competition. Each team was allowed to submit up to two submissions. Eventually, 11 valid submissions from eight different teams were received. Solution names, team members, affiliations, and type of institution (academic or industry) are summarized in Table \ref{tab:partipants_info}. Two teams opted to keep their names and affiliations anonymized. A condensed summary of the details of the approaches (e.g., base architecture, data augmentation, selection of checkpoints) is listed in Table \ref{tab:solutions_info}. Anonymous-2 did not provide a detailed description of their approach. In the following, a brief description of the valid submitted solutions is provided:

\begin{figure*}[ht!]
\begin{center}
\begin{subfigure}[b]{0.47\linewidth}
     \centering
     \includegraphics[width=\linewidth]{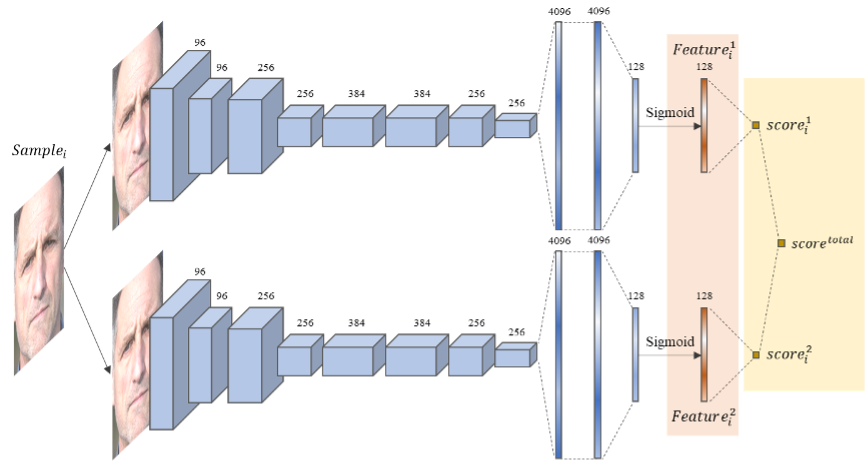}
     \caption{CoDe-Lc}
\end{subfigure}
\begin{subfigure}[b]{0.47\linewidth}
     \centering
     \includegraphics[width=\linewidth]{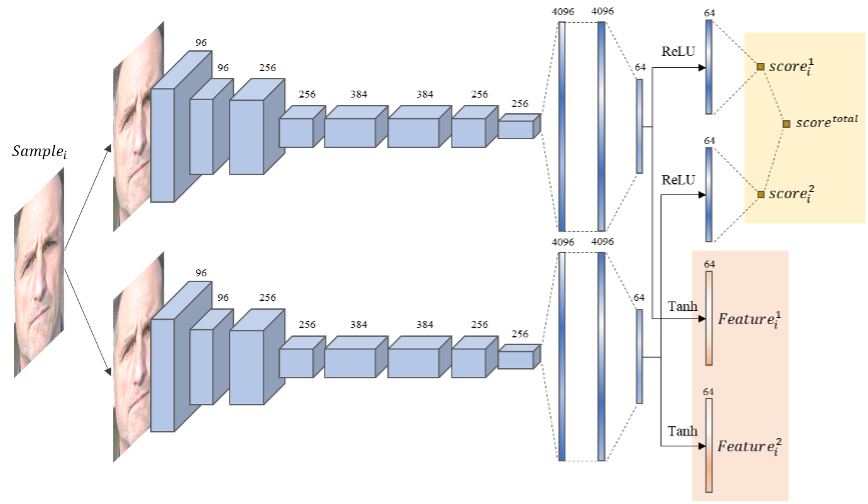}
     \caption{CoDe-Lh}
\end{subfigure}
\caption{The pipeline of (a) CoDe-Lc and (b) CoDe-Lh (rank-3)}
\label{fig:CoDe}
\end{center}
\vspace{-9mm}
\end{figure*}

\textbf{ViT-SIDE B:} The proposed approach involved fine-tuning a pre-trained ViT model architecture \cite{dosovitskiy2020image}, specifically vit-tiny-patch16-224 from the timm library \cite{rw2019timm}, on the provided synthetic data.
To optimize the model's performance, a binary cross-entropy (BCE) loss function was utilized. The Adam \cite{kingma2014adam} optimizer with an initial learning rate of 1e-5 was employed along with step learning rate scheduler, which reduces the learning rate by a factor of 0.7 every 16 epochs until reaching a minimum value of 1e-7.
The model was trained for a minimum of 10 epochs, with the checkpoint exhibiting the minimum loss selected as the final model. 
During training, a weighted sampling strategy was employed to maintain a balanced bona fide (genuine) to attack ratio of 1:1. A batch size of 32 was used, comprising 16 distinct bona fide samples randomly selected. For each bona fide sample, an attack sample with the same identity was paired. The attacks were uniformly sampled from the available sub-attacks.
To enhance the model's robustness, various data augmentations from Albumentation \cite{info11020125} were applied, including JPEG compression, rotation, color jitter, and blurring. These augmentations introduced variations in the training data, enabling the model to handle different types of spoofing attempts.

\textbf{SynFace Co-Former Base (B) and Aug (A):} The proposed Co-Former consists of three Transformer-based branches to extract different-level semantic features, including shallow, normal, and deep semantic features, by using tiny, small, and normal-scale swin-transformer \cite{liu2021swin}. Then, the extracted features are concatenated and processed by two linear layers to cooperatively predict the final results. In addition to Co-Former B using several conventional data augmentation techniques,  Co-Former A is a proposed reflection simulation method to augment the data for imitating the reflective effect caused by the material in practice. Concretely,  Co-Former A randomly selects the coordinate index as the reflection center, and then utilizes a 2-D Gaussian distribution to simulate diffusion reflections. In general, two solutions were given, which were training with the proposed reflection augmentation method (i.e. Co-Former A), and the conventional augmentation methods (horizontal flipping, scaling and rotating, random gamma adjustment, RGB shifting and color jittering) (i.e. Co-Former B), respectively. Two solutions were both trained from scratch, and supervised by the cross-entropy loss function.  The Stochastic Gradient Descent (SGD) optimizer with a momentum of 0.9 and weight decay of 5e-4, and an exponential learning scheduler with a gamma of 0.998 was applied during training. The initial learning rate for training the ResNet models was set to 0.001. The training epoch is 6 to avoiding overfitting and the batch size is set to 14, respectively. Co-Former A serves as a data augmentation method which can be easily incorporated in any other models. The implementation codes and pre-trained models are publicly released \footnote{SynFace Co-Former: \url{https://github.com/Zi-YuanYang/IJCB-SynFacePAD-DIG}} for research reproducibility. Despite two drawbacks associated with Co-Former: a large number of parameters and high computational cost, the performance does not significantly degrade when using only one branch as two branches provided more subtle patterns for PAD decision.

\textbf{CoDe-Lc and CoDe-Lh:}
CoDe-Lc and CoDe-Lh (as shown in Figure \ref{fig:CoDe}) are two ensemble models consisting of dual branches using AlexNet \cite{alexnet} as backbone architecture. Both models were trained from scratch, utilizing a weighted sampling which was performed to ensure a bona fide-attack ratio of 1:1. For CoDe-Lc, the cosine similarity function was employed as the loss function to measure the discrepancy between the feature layers from each branch. Additionally, the Mean Squared Error (MSE) loss was used as an auxiliary metric to evaluate the similarity of features. The BCE loss was computed for the final prediction as well as each branch's prediction. The total loss was calculated as the cumulative sum of all the aforementioned losses. For CoDe-Lh, the cosine similarity was replaced with the hypersphere loss \cite{hploss}. The input images were resized to dimensions of $224 \times 224$, and data augmentation techniques were applied, including random horizontal flipping, scaling and rotating, gamma adjustment, RGB shifting, and color jittering. Moreover, for CoDe-Lh, additional augmentation was introduced by applying random Gaussian blur. The Adam optimizer with a learning rate of 1e-4 and weight decay of 5e-4 was utilized, along with an exponential learning scheduler with a gamma value of 0.998. The batch size during training was set to 128, and the number of training epoch was defined as 200.

\begin{figure*}[htbp]
\centering
\includegraphics[width=0.9\linewidth]{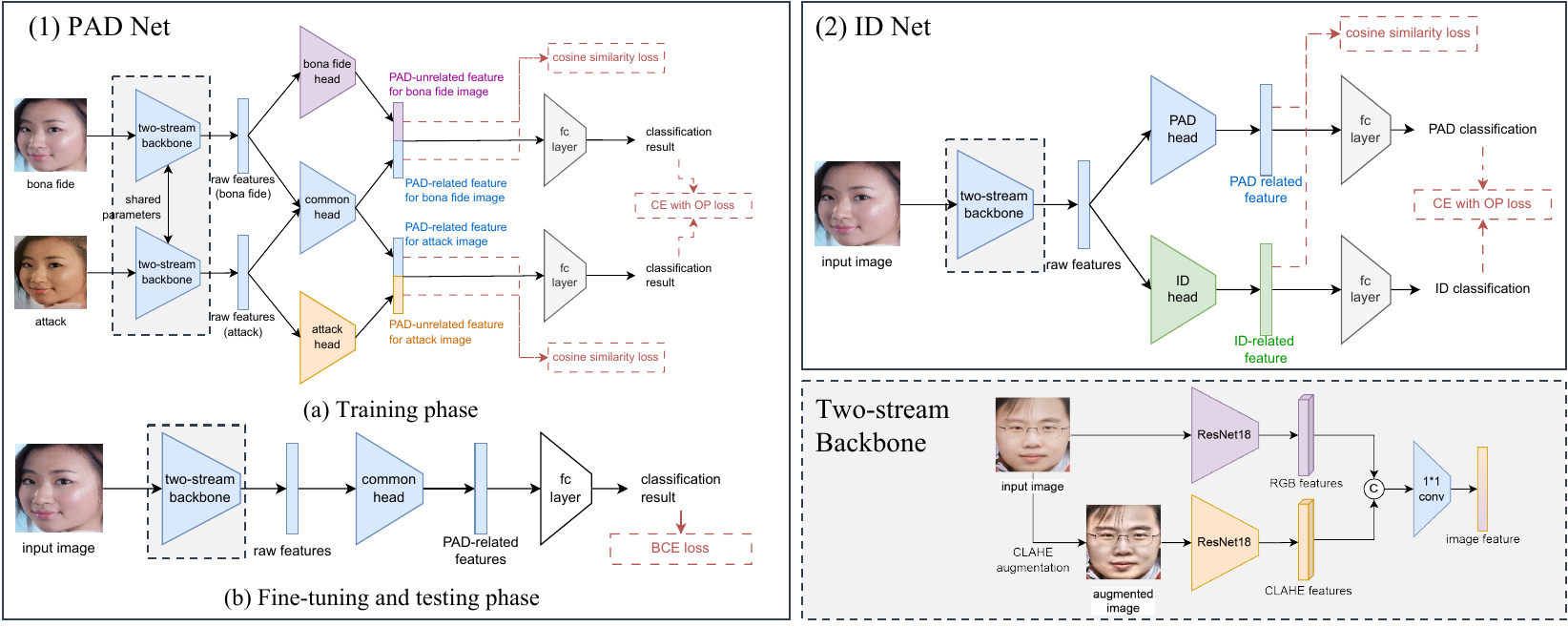}
\caption{The pipeline of OrthPADNet (rank-4) consisting of (1) PAD Net learning PAD task-related features for classification by learning perpendicular feature pairs, and (2) ID Net learning PAD-related but ID-unrelated information. Both PAD Net and ID Net contain a dual-stream backbone to extract features from both the RGB and its CLAHE-augmented version as CLAHE augmentation magnifies the high-frequency details in the images.}
\label{fig:orthPADNet}
\vspace{-5mm}
\end{figure*}
\textbf{OrthPADNet:}
The proposed method, OrthPADNet (as shown in Figure \ref{fig:orthPADNet}), is a fusion of two networks: the PDA net and the ID net. A two-stream backbone is applied to both networks. The two-stream backbone applies two ResNet18 \cite{resnet} to extract features from both the original input and its CLAHE-augmented \cite{clahe} version to fuse a final image feature. The PAD net has a structure similar to \cite{orth}, which extracts two perpendicular features from the raw image feature extracted by the backbone, but only one of them is used for the PAD task. 
The ID net extracts two perpendicular features from the raw features, which are used for PAD and ID classification tasks respectively, and also only the PAD-related part is used for the final decision. Cross entropy loss with orthogonal projection loss \cite{orth-loss} is used for classification, and cosine similarity loss is used for guaranteeing the orthogonality of the features. Apart from the face detection and cropping process provided by the organizers, they applied JPEG compression, rotation, flipping, RGB shift, and eye dropout \cite{facecrop} for data augmentation.

\textbf{hdafvPAD:}
The Fisher Vectors (FV) approach for face PAD \cite{GonzalezSoler-FVGeneralisation-IET-2021} derives a kernel from the parameters of a generative model such as Gaussian Mixture Models (GMM) on $K$-components. In essence, this representation characterises how the distribution of a set of local descriptors, extracted from unknown PAI species, differs from the distribution of known attacks and bona fides, which is previously learned by the generative model. Thus, the most significant properties of the sub-population are summarised. Since image convolution with a suitable filter can effectively quantify frequency variations, local dense-Binarized Statistical Image Features (BSIF) features are used in the approach for image description \cite{GonzalezSoler-denseBSIF-PAD-Biosig-2020}. Given the high correlation among the three RGB color components, these local descriptors are first decorrelated by Principal Component Analysis, thus reducing their size to $d = 64$ components while retaining 95\% of the system variance. Then, the FV representation captures the average first- and second-order statistic differences between the local features and each semantic sub-group previously learned by the GMM. The final FV components are assumed to be independent of each other, allowing the correct use of statistical techniques (e.g., Support Vector Machines) which rely upon assumptions of independence. Overall, the final transformed features are more robust to new samples, which may stem from unknown scenarios and thus differ from the samples used for training.

\textbf{idvcVT:}
The submitted idvcVT is based on the Swin-Transformer architecture \cite{liu2021swin} with a multi-class linear classifier as the final stage. Swin-Transformer builds hierarchical feature maps by merging image patches in deeper layers and has linear computation complexity to input image size due to computation of self-attention only within each local window. Two attack classes were taken into consideration in our case: bona fide and attacks. Input images were transformed according to ImageNet's transformations of RGB images and resized to $256 \times 256 \times 3$ pixels. The model was fine-tuned from ImageNet 1K weights for 50 epochs and results were computed using epoch 10 (best validation performance). Softmax was used on the linear classifier's output to get the score of the bona fide class. 

\textbf{Anonymous-1:}
A ResNet-50 \cite{he2016deep} architecture enhanced with MixStyle for domain generalization \cite{zhou2021mixstyle} and Descriptive Convolutions \cite{Huang_2022_BMVC} with four additional fully connected linear layers for classification. A saliency channel \cite{6115785} was added to the RGB inputs to enhance the detection of certain spoof attacks. The model was supervised using Binary Focal Loss with class weights adjusted to tackle class imbalance. The ADAM optimizer with a betas of $[0.9, 0.999]$ and a weight decay of 5e-4 was used following a cosine annealing scheduler with a period of 20 epochs. The model was trained for 50 epochs with a batch size of 64. Training data is augmented with color jitter, random erasing, and channel normalization.
\vspace{-3mm}
\section{Results and Analysis}

\begin{figure}[ht]
\vspace{-4mm}
\centering
\includegraphics[width=\linewidth]{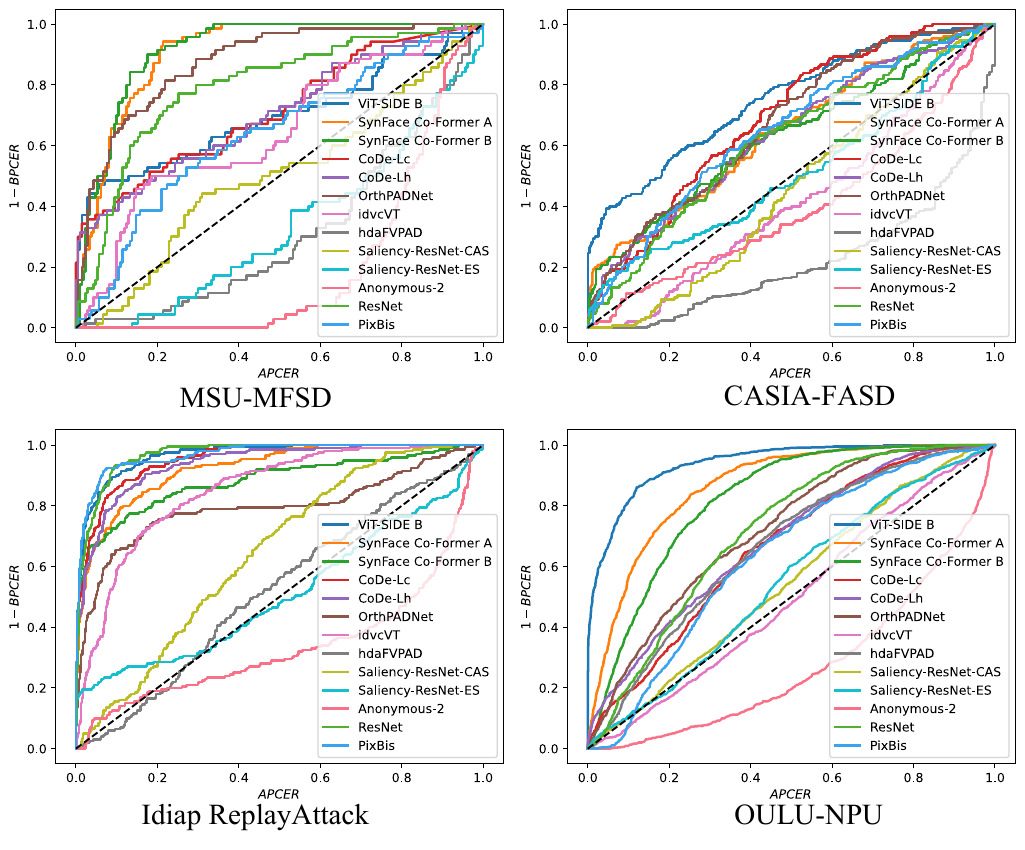}
\caption{ROC curves of 11 submitted solutions and two baseline methods tested on four authentic face PAD benchmarks.}
\label{fig:rocs}
\vspace{-4mm}
\end{figure}

\begin{table*}[ht]
\centering
\resizebox{0.8\textwidth}{!}{
\begin{tabular}{l|cccccc|c}
\hline
\multirow{3}{*}{Solutions} & \multicolumn{6}{c|}{MSU-MFDS \cite{msu_mfs}} & \multirow{5}{*}{Rank} \\ \cline{2-7}
 & \multicolumn{1}{c|}{\multirow{2}{*}{HTER(\%)$\downarrow$}} & \multicolumn{1}{c|}{\multirow{2}{*}{AUC(\%)$\uparrow$}} & \multicolumn{2}{c|}{BPCER(\%)$\downarrow$ @} & \multicolumn{2}{c|}{APCER(\%)$\downarrow$  @ } &  \\ \cline{4-7}
 & \multicolumn{1}{c|}{} & \multicolumn{1}{c|}{} & \multicolumn{1}{c|}{APCER 10\%} & \multicolumn{1}{c|}{APCER 20\%} & \multicolumn{1}{c|}{BPCER 10\%} & \multicolumn{1}{c|}{\textbf{BPCER 20\%}} &  \\ \cline{1-7}
ResNet \cite{synthaspoof} & \multicolumn{1}{c|}{25.48} & \multicolumn{1}{c|}{79.54} & \multicolumn{1}{c|}{57.14} & \multicolumn{1}{c|}{32.86} & \multicolumn{1}{c|}{62.38} & 33.33 &  \\ 
PixBis \cite{synthaspoof} & \multicolumn{1}{c|}{38.33} & \multicolumn{1}{c|}{63.87} & \multicolumn{1}{c|}{84.29} & \multicolumn{1}{c|}{61.43} & \multicolumn{1}{c|}{80.00} & 68.57 &  \\ \hline \hline
SynFace Co-Former B & \multicolumn{1}{c|}{16.67} & \multicolumn{1}{c|}{91.61} & \multicolumn{1}{c|}{32.86} & \multicolumn{1}{c|}{10.00} & \multicolumn{1}{c|}{19.52} & 12.86 & 1 \\ \hline
SynFace Co-Former A & \multicolumn{1}{c|}{18.57} & \multicolumn{1}{c|}{90.76} & \multicolumn{1}{c|}{34.29} & \multicolumn{1}{c|}{12.86} & \multicolumn{1}{c|}{20.48} & 17.62 & 2 \\ \hline
OrthPADNet & \multicolumn{1}{c|}{20.95} & \multicolumn{1}{c|}{87.59} & \multicolumn{1}{c|}{35.71} & \multicolumn{1}{c|}{24.29} & \multicolumn{1}{c|}{33.81} & 23.81 & 3 \\ \hline
CoDe-Lc & \multicolumn{1}{c|}{37.14} & \multicolumn{1}{c|}{71.45} & \multicolumn{1}{c|}{58.57} & \multicolumn{1}{c|}{48.57} & \multicolumn{1}{c|}{68.57} & 57.14 & 4 \\ \hline
CoDe-Lh & \multicolumn{1}{c|}{39.05} & \multicolumn{1}{c|}{70.58} & \multicolumn{1}{c|}{61.43} & \multicolumn{1}{c|}{51.43} & \multicolumn{1}{c|}{74.76} & 60.00 & 5 \\ \hline
idvcVT & \multicolumn{1}{c|}{45.71} & \multicolumn{1}{c|}{64.58} & \multicolumn{1}{c|}{84.29} & \multicolumn{1}{c|}{51.43} & \multicolumn{1}{c|}{84.76} & 61.43 & 6 \\ \hline
ViT-SIDE B & \multicolumn{1}{c|}{36.67} & \multicolumn{1}{c|}{69.78} & \multicolumn{1}{c|}{51.43} & \multicolumn{1}{c|}{47.14} & \multicolumn{1}{c|}{89.52} & 72.38 & 7 \\ \hline
Saliency-ResNet-CAS & \multicolumn{1}{c|}{47.62} & \multicolumn{1}{c|}{50.14} & \multicolumn{1}{c|}{94.29} & \multicolumn{1}{c|}{81.42} & \multicolumn{1}{c|}{90.48} & 82.86 & 8 \\ \hline
Anonymous-2 & \multicolumn{1}{c|}{72.86} & \multicolumn{1}{c|}{19.98} & \multicolumn{1}{c|}{100.00} & \multicolumn{1}{c|}{100.00} & \multicolumn{1}{c|}{68.57} & 90.00 & 9 \\ \hline
hdaFVPAD & \multicolumn{1}{c|}{65.71} & \multicolumn{1}{c|}{31.95} & \multicolumn{1}{c|}{98.57} & \multicolumn{1}{c|}{97.14} & \multicolumn{1}{c|}{96.2} & 92.38 & 10 \\ \hline
Saliency-ResNet-ES & \multicolumn{1}{c|}{58.09} & \multicolumn{1}{c|}{33.49} & \multicolumn{1}{c|}{100} & \multicolumn{1}{c|}{98.71} & \multicolumn{1}{c|}{97.62} & 92.38 & 10 \\ \hline
\end{tabular}}
\caption{The comparative evaluation results of the submitted solutions on MSU-MFSD \cite{msu_mfs} benchmark. The ranking is based on APCER@BPCER=20\%. The top-3 ranked solutions, using multiple branches to capture more sublet PAD patterns, demonstrated an enhanced generalizability.}
\label{tab:msu}
\vspace{-3mm}
\end{table*}

\begin{table*}[ht]
\centering
\resizebox{0.8\textwidth}{!}{
\begin{tabular}{l|cccccc|c}
\hline
 \multirow{3}{*}{Solutions} & \multicolumn{6}{c|}{CASIA-FASD\cite{casia_fas}} & \multirow{5}{*}{Rank} \\ \cline{2-7}
 & \multicolumn{1}{c|}{\multirow{2}{*}{HTER(\%)$\downarrow$}} & \multicolumn{1}{c|}{\multirow{2}{*}{AUC(\%)$\uparrow$}} & \multicolumn{2}{c|}{BPCER (\%)$\downarrow$ @} & \multicolumn{2}{c|}{APCER (\%)$\downarrow$ @ } &  \\ \cline{4-7}
 & \multicolumn{1}{c|}{} & \multicolumn{1}{c|}{} & \multicolumn{1}{c|}{APCER 10\%} & \multicolumn{1}{c|}{APCER 20\%} & \multicolumn{1}{c|}{BPCER 10\%} & \multicolumn{1}{c|}{\textbf{BPCER 20\%}} &  \\ \cline{1-7}
ResNet \cite{synthaspoof}& \multicolumn{1}{c|}{39.22} & \multicolumn{1}{c|}{62.00} & \multicolumn{1}{c|}{84.67} & \multicolumn{1}{c|}{66.67} & \multicolumn{1}{c|}{79.78} & 68.44 &  \\ 
PixBis \cite{synthaspoof} & \multicolumn{1}{c|}{38.44} & \multicolumn{1}{c|}{64.79} & \multicolumn{1}{c|}{79.33} & \multicolumn{1}{c|}{62.67} & \multicolumn{1}{c|}{78.00} & 59.33 &  \\ \hline \hline
ViT-SIDE B & \multicolumn{1}{c|}{33.33} & \multicolumn{1}{c|}{75.21} & \multicolumn{1}{c|}{57.33} & \multicolumn{1}{c|}{45.33} & \multicolumn{1}{c|}{65.56} & 49.11 & 1 \\ \hline
CoDe-Lc & \multicolumn{1}{c|}{37.11} & \multicolumn{1}{c|}{69.08} & \multicolumn{1}{c|}{78.67} & \multicolumn{1}{c|}{66.67} & \multicolumn{1}{c|}{68.44} & 50.44 & 2 \\ \hline
OrthPADNet & \multicolumn{1}{c|}{39.78} & \multicolumn{1}{c|}{67.32} & \multicolumn{1}{c|}{74.67} & \multicolumn{1}{c|}{62.00} & \multicolumn{1}{c|}{66.67} & 56.00 & 3 \\ \hline
CoDe-Lh & \multicolumn{1}{c|}{39.33} & \multicolumn{1}{c|}{63.70} & \multicolumn{1}{c|}{76.67} & \multicolumn{1}{c|}{64.00} & \multicolumn{1}{c|}{79.11} & 63.33 & 4 \\ \hline
SynFace Co-Former A & \multicolumn{1}{c|}{41.11} & \multicolumn{1}{c|}{64.49} & \multicolumn{1}{c|}{72.00} & \multicolumn{1}{c|}{64.00} & \multicolumn{1}{c|}{78.89} & 64.67 & 5 \\ \hline
SynFace Co-Former B & \multicolumn{1}{c|}{40.00} & \multicolumn{1}{c|}{63.05} & \multicolumn{1}{c|}{74.67} & \multicolumn{1}{c|}{61.33} & \multicolumn{1}{c|}{81.56} & 71.78 & 6 \\ \hline
Saliency-ResNet-CAS & \multicolumn{1}{c|}{51.55} & \multicolumn{1}{c|}{45.31} & \multicolumn{1}{c|}{99.33} & \multicolumn{1}{c|}{90.67} & \multicolumn{1}{c|}{86.89} & 79.78 & 7 \\ \hline
Saliency-ResNet-ES & \multicolumn{1}{c|}{51.11} & \multicolumn{1}{c|}{51.09} & \multicolumn{1}{c|}{80.00} & \multicolumn{1}{c|}{74.00} & \multicolumn{1}{c|}{87.78} & 83.11 & 8 \\ \hline
idvcVT & \multicolumn{1}{c|}{56.44} & \multicolumn{1}{c|}{41.82} & \multicolumn{1}{c|}{98.67} & \multicolumn{1}{c|}{89.33} & \multicolumn{1}{c|}{91.56} & 85.33 & 9 \\ \hline
Anonymous-2 & \multicolumn{1}{c|}{60.22} & \multicolumn{1}{c|}{40.04} & \multicolumn{1}{c|}{90.67} & \multicolumn{1}{c|}{84.00} & \multicolumn{1}{c|}{94.22} & 89.33 & 10 \\ \hline
hdaFVPAD & \multicolumn{1}{c|}{71.33} & \multicolumn{1}{c|}{21.95} & \multicolumn{1}{c|}{100.00} & \multicolumn{1}{c|}{97.56} & \multicolumn{1}{c|}{99.33} & 97.33 & 11 \\ \hline
\end{tabular}}
\caption{The comparative evaluation results of the submitted solutions on CASIA-FASD \cite{casia_fas} benchmark. The submitted solutions generalized not well on CAISA-FASD, in comparison to on the other three benchmarks. The possible reason is that CASIA-FASD contains a printed photo attack sample with eye region cut out, which is not present in the synthetic training data.}
\label{tab:casia}
\vspace{-5mm}
\end{table*}

\begin{table*}[ht]
\centering
\resizebox{0.8\textwidth}{!}{
\begin{tabular}{l|cccccc|c}
\hline
\multirow{3}{*}{Solutions} & \multicolumn{6}{c|}{Idaip ReplayAttack\cite{replay_attack}} & \multirow{5}{*}{Rank} \\ \cline{2-7}
 & \multicolumn{1}{c|}{\multirow{2}{*}{HTER(\%)$\downarrow$}} & \multicolumn{1}{c|}{\multirow{2}{*}{AUC(\%)$\uparrow$}} & \multicolumn{2}{c|}{BPCER (\%)$\downarrow$ @} & \multicolumn{2}{c|}{APCER (\%)$\downarrow$ @ } &  \\ \cline{4-7}
 & \multicolumn{1}{c|}{} & \multicolumn{1}{c|}{} & \multicolumn{1}{c|}{APCER 10\%} & \multicolumn{1}{c|}{APCER 20\%} & \multicolumn{1}{c|}{BPCER 10\%} & \multicolumn{1}{c|}{\textbf{BPCER 20\%}} &  \\ \cline{1-7}
ResNet \cite{synthaspoof} & \multicolumn{1}{c|}{8.90} & \multicolumn{1}{c|}{96.96} & \multicolumn{1}{c|}{7.00} & \multicolumn{1}{c|}{2.50} & \multicolumn{1}{c|}{8.50} & 5.10 &  \\ 
PixBis \cite{synthaspoof} & \multicolumn{1}{c|}{7.50} & \multicolumn{1}{c|}{96.88} & \multicolumn{1}{c|}{7.50} & \multicolumn{1}{c|}{5.50} & \multicolumn{1}{c|}{6.50} & 3.00 &  \\ \hline\hline
ViT-SIDE B & \multicolumn{1}{c|}{9.80} & \multicolumn{1}{c|}{96.67} & \multicolumn{1}{c|}{10.50} & \multicolumn{1}{c|}{3.50} & \multicolumn{1}{c|}{10.50} & 4.40 & 1 \\ \hline
CoDe-Lc & \multicolumn{1}{c|}{12.10} & \multicolumn{1}{c|}{95.31} & \multicolumn{1}{c|}{14.50} & \multicolumn{1}{c|}{7.00} & \multicolumn{1}{c|}{15.70} & 6.60 & 2 \\ \hline
CoDe-Lh & \multicolumn{1}{c|}{13.90} & \multicolumn{1}{c|}{93.84} & \multicolumn{1}{c|}{18.50} & \multicolumn{1}{c|}{9.00} & \multicolumn{1}{c|}{16.90} & 9.40 & 3 \\ \hline
SynFace Co-Former A & \multicolumn{1}{c|}{16.30} & \multicolumn{1}{c|}{92.31} & \multicolumn{1}{c|}{23.00} & \multicolumn{1}{c|}{14.50} & \multicolumn{1}{c|}{24.20} & 13.20 & 4 \\ \hline
SynFace Co-Former B & \multicolumn{1}{c|}{18.80} & \multicolumn{1}{c|}{88.20} & \multicolumn{1}{c|}{27.50} & \multicolumn{1}{c|}{19.00} & \multicolumn{1}{c|}{41.90} & 18.70 & 5 \\ \hline
idvcVT & \multicolumn{1}{c|}{23.10} & \multicolumn{1}{c|}{85.08} & \multicolumn{1}{c|}{42.00} & \multicolumn{1}{c|}{25.50} & \multicolumn{1}{c|}{40.40} & 25.40 & 6 \\ \hline
OrthPADNet & \multicolumn{1}{c|}{23.70} & \multicolumn{1}{c|}{79.55} & \multicolumn{1}{c|}{35.50} & \multicolumn{1}{c|}{25.50} & \multicolumn{1}{c|}{78.00} & 57.30 & 7 \\ \hline
Saliency-ResNet-CAS & \multicolumn{1}{c|}{40.80} & \multicolumn{1}{c|}{64.48} & \multicolumn{1}{c|}{85.00} & \multicolumn{1}{c|}{73.00} & \multicolumn{1}{c|}{67.10} & 57.90 & 8 \\ \hline
hdaFVPAD & \multicolumn{1}{c|}{47.80} & \multicolumn{1}{c|}{52.10} & \multicolumn{1}{c|}{94.00} & \multicolumn{1}{c|}{83.00} & \multicolumn{1}{c|}{89.70} & 74.70 & 9 \\ \hline
Saliency-ResNet-ES & \multicolumn{1}{c|}{50.70} & \multicolumn{1}{c|}{51.25} & \multicolumn{1}{c|}{75.50} & \multicolumn{1}{c|}{72.00} & \multicolumn{1}{c|}{94.10} & 86.20 & 10 \\ \hline
Anonymous-2 & \multicolumn{1}{c|}{64.00} & \multicolumn{1}{c|}{34.05} & \multicolumn{1}{c|}{87.50} & \multicolumn{1}{c|}{81.00} & \multicolumn{1}{c|}{96.50} & 94.50 & 11 \\ \hline
\end{tabular}}
\caption{The comparative evaluation results of the submitted solutions on Idiap ReplayAttack \cite{replay_attack} benchmark. Most submitted solutions obtained the best performance on this benchmarks which may be benefit from a diverse replay attack in training set. }
\label{tab:replayattack}
\vspace{-3mm}
\end{table*}

\begin{table*}[ht]
\centering
\resizebox{0.8\textwidth}{!}{
\begin{tabular}{l|cccccc|c}
\hline
\multirow{3}{*}{Solutions} & \multicolumn{6}{c|}{OULU-NPU\cite{oulu_npu}} & \multirow{5}{*}{Rank} \\ \cline{2-7}
 & \multicolumn{1}{c|}{\multirow{2}{*}{HTER(\%)$\downarrow$}} & \multicolumn{1}{c|}{\multirow{2}{*}{AUC(\%)$\uparrow$}} & \multicolumn{2}{c|}{BPCER (\%)$\downarrow$ @} & \multicolumn{2}{c|}{APCER (\%)$\downarrow$ @ } &  \\ \cline{4-7}
 & \multicolumn{1}{c|}{} & \multicolumn{1}{c|}{} & \multicolumn{1}{c|}{APCER 10\%} & \multicolumn{1}{c|}{APCER 20\%} & \multicolumn{1}{c|}{BPCER 10\%} & \multicolumn{1}{c|}{\textbf{BPCER 20\%}} &  \\ \cline{1-7}
ResNet \cite{synthaspoof} & \multicolumn{1}{c|}{31.48} & \multicolumn{1}{c|}{71.48} & \multicolumn{1}{c|}{79.80} & \multicolumn{1}{c|}{59.70} & \multicolumn{1}{c|}{57.07} & 45.56 &  \\ 
PixBis \cite{synthaspoof}& \multicolumn{1}{c|}{35.77} & \multicolumn{1}{c|}{67.71} & \multicolumn{1}{c|}{86.57} & \multicolumn{1}{c|}{64.75} & \multicolumn{1}{c|}{65.56} & 49.60 &  \\ \hline \hline
ViT-SIDE B & \multicolumn{1}{c|}{13.26} & \multicolumn{1}{c|}{94.04} & \multicolumn{1}{c|}{18.79} & \multicolumn{1}{c|}{8.69} & \multicolumn{1}{c|}{17.68} & 9.14 & 1 \\ \hline
SynFace Co-Former A & \multicolumn{1}{c|}{21.67} & \multicolumn{1}{c|}{86.44} & \multicolumn{1}{c|}{43.63} & \multicolumn{1}{c|}{23.74} & \multicolumn{1}{c|}{33.08} & 22.88 & 2 \\ \hline
SynFace Co-Former B & \multicolumn{1}{c|}{25.35} & \multicolumn{1}{c|}{82.02} & \multicolumn{1}{c|}{61.11} & \multicolumn{1}{c|}{34.24} & \multicolumn{1}{c|}{40.23} & 29.34 & 3 \\ \hline
OrthPADNet & \multicolumn{1}{c|}{34.92} & \multicolumn{1}{c|}{71.69} & \multicolumn{1}{c|}{73.13} & \multicolumn{1}{c|}{54.55} & \multicolumn{1}{c|}{60.76} & 49.14 & 4 \\ \hline
hdaFVPAD & \multicolumn{1}{c|}{37.89} & \multicolumn{1}{c|}{66.49} & \multicolumn{1}{c|}{81.82} & \multicolumn{1}{c|}{62.53} & \multicolumn{1}{c|}{70.10} & 54.89 & 5 \\ \hline
CoDe-Lh & \multicolumn{1}{c|}{38.11} & \multicolumn{1}{c|}{68.33} & \multicolumn{1}{c|}{75.66} & \multicolumn{1}{c|}{58.59} & \multicolumn{1}{c|}{69.09} & 55.73 & 6 \\ \hline
CoDe-Lc & \multicolumn{1}{c|}{37.58} & \multicolumn{1}{c|}{66.30} & \multicolumn{1}{c|}{81.21} & \multicolumn{1}{c|}{66.46} & \multicolumn{1}{c|}{68.46} & 56.94 & 7 \\ \hline
Saliency-ResNet-ES & \multicolumn{1}{c|}{46.03} & \multicolumn{1}{c|}{54.7} & \multicolumn{1}{c|}{89.29} & \multicolumn{1}{c|}{81.11} & \multicolumn{1}{c|}{83.61} & 70.71 & 8 \\ \hline
Saliency-ResNet-CAS & \multicolumn{1}{c|}{47.55} & \multicolumn{1}{c|}{54.4} & \multicolumn{1}{c|}{89.89} & \multicolumn{1}{c|}{77.68} & \multicolumn{1}{c|}{83.31} & 71.67 & 9 \\ \hline
idvcVT & \multicolumn{1}{c|}{51.09} & \multicolumn{1}{c|}{49.68} & \multicolumn{1}{c|}{93.03} & \multicolumn{1}{c|}{83.13} & \multicolumn{1}{c|}{86.21} & 76.59 & 10 \\ \hline
Anonymous-2 & \multicolumn{1}{c|}{66.39} & \multicolumn{1}{c|}{27.86} & \multicolumn{1}{c|}{99.09} & \multicolumn{1}{c|}{95.35} & \multicolumn{1}{c|}{98.06} & 95.61 & 11 \\ \hline
\end{tabular}}
\caption{The comparative evaluation results of the submitted solutions on OULU-NPU \cite{oulu_npu} benchmark. The top-3 ranked solutions obtained a superior performance. In addition to their innovative designs, this may be related to two shared factors 1) the base networks of these solutions are self-attention based transformers, 2) extensive augmentation techniques. }
\label{tab:oulu}
\vspace{-4mm}
\end{table*}

\begin{table}[ht]
\centering
\resizebox{0.45\textwidth}{!}{
\begin{tabular}{l|l|cccccc}
\hline
\multirow{2}{*}{Teams} & \multirow{2}{*}{Solutions} & \multicolumn{6}{c}{Ranks} \\ \cline{3-8} 
 &  & \multicolumn{1}{c|}{M} & \multicolumn{1}{c|}{C} & \multicolumn{1}{c|}{I} & \multicolumn{1}{c|}{O} & \multicolumn{1}{c|}{Avg} & Final rank \\ \hline \hline
ID R\&D Inc & ViT-SIDE B & \multicolumn{1}{c|}{7} & \multicolumn{1}{c|}{1} & \multicolumn{1}{c|}{1} & \multicolumn{1}{c|}{1} & \multicolumn{1}{c|}{\textbf{2.50}} & 1 \\ \hline
\multirow{2}{*}{SCU-DIG} & SynFace Co-Former A & \multicolumn{1}{c|}{2} & \multicolumn{1}{c|}{5} & \multicolumn{1}{c|}{4} & \multicolumn{1}{c|}{2} & \multicolumn{1}{c|}{\textbf{3.25}} & \multirow{2}{*}{2} \\ 
 & SynFace Co-Former B & \multicolumn{1}{c|}{1} & \multicolumn{1}{c|}{6} & \multicolumn{1}{c|}{5} & \multicolumn{1}{c|}{3} & \multicolumn{1}{c|}{3.75} &  \\ \hline
\multirow{2}{*}{HIT} & CoDe-Lc & \multicolumn{1}{c|}{4} & \multicolumn{1}{c|}{2} & \multicolumn{1}{c|}{2} & \multicolumn{1}{c|}{7} & \multicolumn{1}{c|}{\textbf{3.75}} & \multirow{2}{*}{3} \\ 
 & CoDe-Lh & \multicolumn{1}{c|}{5} & \multicolumn{1}{c|}{4} & \multicolumn{1}{c|}{3} & \multicolumn{1}{c|}{6} & \multicolumn{1}{c|}{4.50} &  \\ \hline
BUCEA & OrthPADNet & \multicolumn{1}{c|}{3} & \multicolumn{1}{c|}{3} & \multicolumn{1}{c|}{7} & \multicolumn{1}{c|}{4} & \multicolumn{1}{c|}{\textbf{4.25}} & 4 \\ \hline
idvc & idvcVT & \multicolumn{1}{c|}{6} & \multicolumn{1}{c|}{9} & \multicolumn{1}{c|}{6} & \multicolumn{1}{c|}{10} & \multicolumn{1}{c|}{\textbf{7.75}} & {5} \\ \hline 
\multirow{2}{*}{Anonymous-1} & Saliency-ResNet-CAS & \multicolumn{1}{c|}{8} & \multicolumn{1}{c|}{7} & \multicolumn{1}{c|}{8} & \multicolumn{1}{c|}{9} & \multicolumn{1}{c|}{\textbf{8.00}} & \multirow{2}{*}{6} \\ 
& Saliency-ResNet-ES & \multicolumn{1}{c|}{10} & \multicolumn{1}{c|}{8} & \multicolumn{1}{c|}{10} & \multicolumn{1}{c|}{8} & \multicolumn{1}{c|}{9.00} &  \\ \hline
hda & hdaFVPAD & \multicolumn{1}{c|}{10} & \multicolumn{1}{c|}{11} & \multicolumn{1}{c|}{9} & \multicolumn{1}{c|}{5} & \multicolumn{1}{c|}{\textbf{8.75}} & 7 \\ \hline
Anonymous-2 & Anonymous-2 & \multicolumn{1}{c|}{9} & \multicolumn{1}{c|}{10} & \multicolumn{1}{c|}{11} & \multicolumn{1}{c|}{11} & \multicolumn{1}{c|}{\textbf{10.25}} & 8 \\ \hline
\end{tabular}}
\caption{The final ranking of the participating teams based on the average ranking of their best performance of the submitted solutions on four benchmarks. The bold number is the best performance of each team. }
\label{tab:final_rank}
\vspace{-5mm}
\end{table}

This section presents the evaluation results of the submitted solutions and the baselines on four authentic face PAD datasets in terms of the metrics introduced in Section \ref{sec:comp_info}. Note that APCER at BPCER of 20\% is used for ranking solutions and the final ranking is based on the average ranking on all benchmarks. As shown in Table \ref{tab:final_rank}, Team ID R\&D Inc with solution ViT-SIDE B achieved the top-1 rank.
\vspace{-2mm}
\subsection{Analysis}

\textbf{MSU-MFSD:} Table \ref{tab:msu} presents the results of all submitted solutions and the two baseline methods on MSU-MFSD benchmarks. We observe that (1) three solutions outperformed baseline ResNet and six solutions outperformed PixBis in terms of APCER@BPCER=20\%. (2) SynFace Co-Former B and A achieved the lowest two error rates. (3) Multiple branches help in capturing fine-grained and subtle PAD patterns, as evidenced by the top-3 ranked solutions.

\textbf{CASIA-FASD:} Table \ref{tab:casia} presents the results of all submitted solutions and two baseline methods on CASIA-FASD benchmarks. We observe that (1) five solutions outperformed baseline methods. (2) ViT-SIDE B achieved the best performance, followed by CoDe-Lc and OrthPADNet. (3) all solutions achieved APCER higher than 48\% at BPCER of 20\%, indicating that CASIA-FASD is more challenging than the other three benchmarks. This might be attributed to that CASIA-FASD includes a photo attack with eye region cut out, which is not seen in the training data.

\textbf{Idiap ReplayAttack:} Table \ref{tab:replayattack} presents the results of all submitted solutions and two baseline methods on the Idiap ReplayAttack benchmark. We observe that (1) Most submitted solutions achieved their best performance on this benchmark, maybe due to a diverse range of replay attacks (more than print attacks) in the training data of SynthASpoof \cite{synthaspoof}. (2) The best performance in terms of APCER@BPCER=20\% is achieved by the ViT-SIDE B solution, where the achieved APCER is 4.40\%. 

\textbf{OULU-NPU:} Table \ref{tab:oulu} presents the results of all submitted solutions and two baseline methods on OULU-NPU benchmark. We observe that (1) ViT-SIDE B achieved superior performance compared to all solutions, with an APCER of 9.14\% at 20\% BPCER and the rank 2 solution SynFace Co-Former A achieved an APCER of 22.88\% at 20\% BPCER. (2) The top 3 ranked solutions employed transformer as the base network, suggesting the relatively higher generalizability of self-attention based transformer.

Figure \ref{fig:rocs} presents the achieved performance in terms of ROC curves on each benchmark by 11 submitted solutions and two baseline methods. A consistent observation can be made: 1) Most of the submitted solutions achieved better performance on MSU-MFSD, Idiap ReplayAttack, and OULU-NPU benchmarks, in comparison to CASIA-FASD. 2) Most of the presented solutions are very competitive and achieve better performance compared to the two baseline models. 
The results of Anonymous-2 solutions indicate a strong over-fitting of the model. However, it is hard to make a specific analysis without information on the submitted model.
Notely, all models were limited to training on the synthetic data without any access to authentic faces and the best checkpoint is mostly determined by the fixed epochs or the training loss (details in Table \ref{tab:solutions_info} and descriptions in Section \ref{sec:submitted_solutions}). This further proved the feasibility of using synthetic data for developing face PADs. In addition, the possible domain gaps between authentic and synthetic data can be targeted in the future by including more attack types and formulate such problem in cross-domain.

\vspace{-3mm}
\subsection{Comparison and Final Ranking}

Table \ref{tab:final_rank} presents the final ranking based on the average rank achieved on the four authentic face PAD benchmarks. From the ranking outcome, we made the following general observations: \textbf{(1)} Solutions using transformer-based architecture as the base network generally exhibited higher PA detectability compared to CNNs. For example, ViT-SIDE B based on ViT-Tiny and SynFace Co-Former based on Swin-Transformer ranked first and second, respectively. \textbf{(2)} The incorporation of diverse data augmentation techniques helped in enhancing the generalizability of PADs. This can be observed when comparing SynFace Co-Former and idvcVT, both of which employ Swin-Transformer as the base architecture. \textbf{(3)} Using multiple branches or models contributed to an accurate and generalized PAD decision. \textbf{(4)} Deep-learning based solutions obtained outperformed hand-crafted feature-based methods in most cases.

In the final ranking, Team ID R\&D Inc with solution ViT-SIDE B won the competition with an average rank of 2.50, the second place was achieved by the SCU-DIG team with SynFace Co-Former A model (average rank 3.25), and the third place is obtained by Team HIT with CoDe-Lc model (average rank 3.75), as detailed in Table \ref{tab:final_rank}.

\vspace{-3mm}
\section{Conclusion}
\vspace{-1mm}
In this paper, we summarized the results and observations of the SynFacePAD 2023: Competition on Face Presentation Attack Detection Based on Privacy-aware Synthetic Training Data. In total, 14 teams registered for participation, and eight of them submitted 11 valid submissions to address the challenges of face PAD while considering privacy and legal concerns associated with authentic development data. 
The evaluation of the submitted solutions was conducted on four publicly available authentic face PAD benchmarks. The competition showcased various innovative approaches, resulting in improved performance compared to the considered two baseline methods. The enhanced PAD performance demonstrated the feasibility of using synthetic data and highlighted the potential of synthetic data in the development of face PAD systems.
\vspace{-4mm}
\paragraph{Acknowledgment:}
This research work has been funded by the German Federal Ministry of Education and Research and the Hessen State Ministry for Higher Education, Research and the Arts within their joint support of the National Research Center for Applied Cybersecurity ATHENE. J.F. is supported by project BBforTAI (PID2021-127641OB-I00MICINN/FEDER).

\clearpage

{\small
\bibliographystyle{ieee}
\bibliography{egbib}
}

\end{document}